\documentclass[letterpaper, 10 pt, conference]{ieeeconf}

\IEEEoverridecommandlockouts
\usepackage{graphicx}
\usepackage{amsmath,amssymb,amsfonts}
\usepackage{booktabs}
\usepackage{multirow}
\usepackage{url}
\usepackage{cite}
\usepackage{xcolor}
\usepackage{capt-of}
\usepackage{placeins}
\usepackage{algorithm}
\usepackage{algorithmic}

\usepackage{hyperref}
\usepackage{arydshln}
\usepackage{pifont}
\newcommand{\cmark}{\textcolor{green!50!black}{\ding{51}}}
\newcommand{\xmark}{\textcolor{red!80!black}{\ding{55}}}

\usepackage{tikz}
\usetikzlibrary{calc}

\renewcommand{\footnoterule}{%
    \kern -3pt
    \hrule width 0.4\columnwidth height 0.4pt
    \kern 2.6pt
}

\title{FoLD: Force-Informed Learning for Dexterous \\ Articulated Object Manipulation}

\author{%
    \authorblockN{Haowei Shen$^{1}$, Tingai Li$^{1}$, Yumeng Liu$^{1,*}$, Wenyuan Guang$^{2}$,\\
    Xuanze Yang$^{1}$, Qing Fang$^{1}$, Kai Xu$^{2,3}$, Ligang Liu$^{1}$, Ruizhen Hu$^{4,*}$}
    \authorblockA{$^{1}$University of Science and Technology of China\\
    $^{2}$Institute of AI for Industries, Chinese Academy of Sciences\\
    $^{3}$Jiangsu Key Laboratory of AI for Industries\\
    $^{4}$Shenzhen University}
    \thanks{$^{*}$Corresponding authors: Yumeng Liu (\href{mailto:lym29@mail.ustc.edu.cn}{\texttt{lym29@mail.ustc.edu.cn}}) and Ruizhen Hu (\href{mailto:ruizhen.hu@gmail.com}{\texttt{ruizhen.hu@gmail.com}}).}
}

\hypersetup{
    hidelinks,
    pdftitle={FoLD: Force-Informed Learning for Dexterous Articulated Object Manipulation},
    pdfauthor={Haowei Shen, Tingai Li, Yumeng Liu, Wenyuan Guang, Xuanze Yang, Qing Fang, Kai Xu, Ligang Liu, Ruizhen Hu},
    pdfsubject={Research preprint},
    pdfkeywords={Dexterous Manipulation, Learning from Demonstration, Imitation Learning}
}

\IEEEaftertitletext{%
    \begin{minipage}{\textwidth}
        \centering
        \vspace{-3em}
        \includegraphics[width=\linewidth]{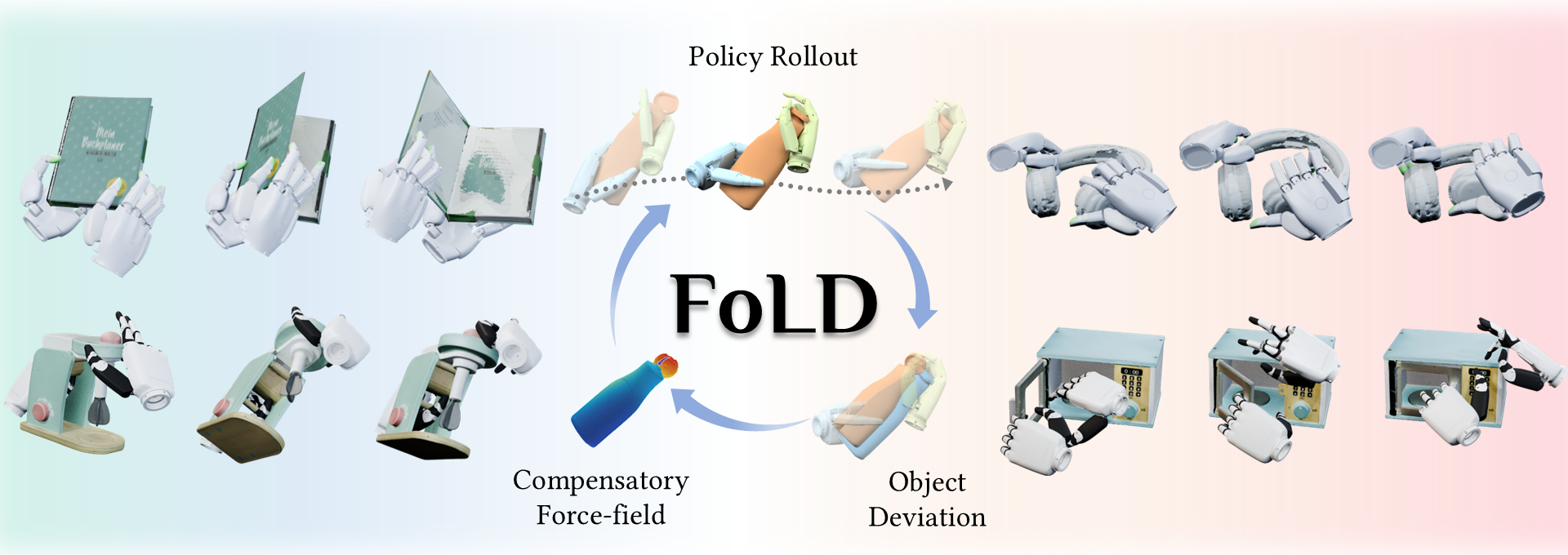}
        \vspace{-1em}
        \captionof{figure}{We present FoLD, a force-informed learning framework for dexterous manipulation of articulated objects from human demonstrations. FoLD uses object-motion deviations to estimate virtual compensatory force fields that inform corrective hand actions, bringing object motion closer to the demonstration.}
        \label{fig:teaser}
    \end{minipage}
}

\begin{document}
\maketitle
\thispagestyle{plain}
\pagestyle{plain}

\begin{abstract}
Transferring human demonstrations to dexterous robots remains challenging because differences in hand morphology and contact dynamics often cause retargeted motions to fail at producing the intended object behavior. We present \textbf{FoLD}, a framework for learning dexterous manipulation of articulated objects through explicit force guidance. FoLD compute compensatory force fields from human demonstrations together with the robot's current interaction state, yielding a force prior that promotes the demonstrated object motion. This force prior informs a residual policy that adapts retargeted hand motions to the contact requirements of the task. We evaluate FoLD on a public benchmark for articulated object manipulation, where it consistently outperforms state-of-the-art baselines across tasks and embodiments. We further validate FoLD on real dexterous robot platforms, demonstrating successful transfer of human manipulation skills to robot execution. Here is the link of our project page: \url{https://gghgghgghgg.github.io/FoLD-project-page/}.
\end{abstract}

\section{Introduction}
\label{sec:introduction}

Dexterous hands offer a general interface for interacting with the physical world. Their rich kinematic structure enables robots to reproduce the coordinated hand configurations and finger motions involved in everyday manipulation~\cite{okamura2000overview}. An important application is the manipulation of articulated objects, whose operation requires coordinated hand movements to change the relative configuration of connected object parts. Yet learning to manipulate articulated objects is challenging due to the hand's high-dimensional action space and the motion constraints imposed by object joints. Human demonstrations can reduce the exploration burden of learning dexterous manipulation~\cite{rajeswaran2018dapg}. For articulated objects, they capture how hand movements accompany changes in object articulation throughout a task. Learning from such demonstrations offers a scalable path toward dexterous skills for operating articulated objects.

Despite this promise, transferring human demonstrations to dexterous robots remains challenging. Human and robot hands differ in morphology, joint structure, and motion range, so kinematic retargeting can only provide an approximate motion reference. For articulated objects, the constraints imposed by object joints determine how contact interactions translate into motion of individual parts. Small discrepancies in a retargeted hand pose can therefore alter where contact occurs and whether the interaction produces the intended articulation. Early interaction errors can prevent the policy from progressing through the demonstrated sequence, thereby compromising the realization of the intended manipulation function.

Early human-to-robot retargeting methods transfer human hand motions to robot hands by matching joint configurations or task-space keypoints~\cite{handa2020dexpilot,sivakumar2022robotic,qin2023anyteleop}. While effective for teleoperation and motion generation, these kinematics-centric objectives do not account for object dynamics or manipulation outcomes. Recent approaches therefore leverage reconstructed hand--object trajectories to optimize physically feasible motions or control policies~\cite{qin2021dexmv,liu2024quasisim,liu2025dextrack,li2025maniptrans,zhao2026dexmachina,zhu2026chord}. DexMachina~\cite{zhao2026dexmachina} uses virtual object-level controllers to stabilize the reference object trajectory during early policy optimization. However, because this assistance acts directly on the object, it does not specify how the robot hand should coordinate contact forces to sustain the motion after the controller decays. Moving closer to the contact interface, CHORD~\cite{zhu2026chord} compares the wrench directions supported by human and robot contact configurations and uses this representation to guide policy optimization. Although the policy observes the current contact locations and force directions, the demonstration-derived wrench-support representation serves as evaluative supervision rather than state-conditioned mechanical guidance for action generation. Such supervision evaluates the quality of the resulting interaction, but leaves the desired force adjustments implicit when the policy selects an action. This motivates providing the policy with a description of how contact forces could correct the current object-motion deviation, making mechanical guidance available during action prediction.

Our key insight is to express this guidance as a virtual compensatory force field that informs residual action prediction. The resulting residual actions adjust the robot hand poses and contacts so that the object motion more closely follows the demonstration. We introduce \emph{FoLD} (Force-Informed Learning for Dexterous Articulated Object Manipulation), a force-informed learning framework for dexterous manipulation from human demonstrations. Starting from reference actions produced by a hand imitator, FoLD estimates a compensatory force field from object-motion deviations and the contact regions available to the robot hands. This field describes where and how forces could compensate for the deviation from the demonstrated motion. We encode the field into a latent representation and provide it to a residual policy as an input for predicting corrections to the imitator's reference actions. Adding these residual actions adjusts the hand poses and resulting contacts, allowing the policy to learn how to realize the desired object motion through the robot's own interaction forces.

We validate FoLD on bimanual articulated object manipulation using human demonstrations from the public ARCTIC dataset~\cite{fan2023arctic}. Across the evaluated tasks, FoLD achieves state-of-the-art performance over existing baselines. Ablation studies further confirm the importance of compensatory force-field conditioning for effective policy learning. Moreover, by training and evaluating FoLD on the same demonstrated tasks with different robot hands, we show that its effectiveness is consistent across embodiments. Together, these results highlight the effectiveness of FoLD for dexterous articulated object manipulation.

Our main contributions are:
\begin{itemize}
    \item We introduce FoLD, a force-informed learning framework for dexterous manipulation of articulated objects from human demonstrations. FoLD learns to use virtual compensatory force fields to correct retargeted hand motions, bringing the resulting object motion closer to the demonstration.
    \item We formulate a constrained inverse-mechanics problem that infers compensatory force fields from human motion demonstrations and the current robot interaction state, providing a physically plausible force prior to inform residual policy learning.
    \item We achieve state-of-the-art performance on the evaluated ARCTIC manipulation tasks. Through extensive experiments across dexterous hand embodiments and ablation studies, we demonstrate that force-informed learning substantially improves the acquisition of dexterous manipulation skills.
\end{itemize}

\section{Related Work}
\label{sec:related_work}

\subsection{Learning Manipulation from Human Demonstrations}
Human demonstrations capture the coordinated hand motions underlying dexterous manipulation and provide supervision for robot learning. Teleoperation-based imitation collects robot actions directly but requires access to the target hardware~\cite{arunachalam2023holodex,qin2023anyteleop,liu2024realdex,arunachalam2023dime}. Kinesthetic teaching collects demonstrations with tactile feedback through direct guidance of the robot hand~\cite{zhang2025kinedex}. Retargeting bridges different hand morphologies through relative fingertip positions~\cite{wang2004relative}, object-based synergy mappings~\cite{gioioso2013mapping}, or shared low-dimensional pose spaces~\cite{meeker2018intuitive}. For offline demonstrations, visual reconstruction and retargeting turn human videos into robot training references~\cite{qin2021dexmv,liu2026egoengine}, reducing dependence on robot-specific data collection. However, matching motion alone does not ensure that the robot establishes the contacts needed to manipulate the object. Task-oriented retargeting therefore incorporates manipulation objectives~\cite{antotsiou2018task}, while physics-based methods optimize or sample trajectories under simulated dynamics~\cite{liu2024quasisim,pan2025spider}. Contact-aware optimization recovers contact points and forces from visual demonstrations~\cite{zhu2023difflfd}. Demonstration-guided reinforcement learning uses motion and task rewards to acquire feedback policies~\cite{chen2024objdex,liu2025dextrack,lum2025human2sim2robot}. Adaptive exploration scopes accommodate noisy references and embodiment differences~\cite{xu2025dexplore}. Residual learning further corrects estimated hand poses~\cite{garciahernando2020physics} or adjusts a learned motion prior during interaction~\cite{li2025maniptrans}. FoLD complements these approaches by estimating a compensatory force field from the current object-motion deviation and conditioning a residual policy on this field. The field specifies how forces could correct the deviation, guiding adjustments to the hand-motion prior.

\subsection{Manipulation of Articulated Objects}
Articulated objects such as boxes, laptops, and scissors are common in everyday manipulation and require coordinated control of connected parts. Visual affordance approaches predict contact locations and action or trajectory proposals for moving articulated parts~\cite{mo2021where2act,wu2022vatmart}. These predictions guide part actuation, but do not directly describe how two dexterous hands should coordinate contacts to support and manipulate a moving object. Human interaction datasets provide examples of such coordination: ARCTIC focuses on hinged-object manipulation~\cite{fan2023arctic}, while OakInk2 includes articulated objects within broader bimanual tasks~\cite{zhan2024oakink2}. These datasets provide valuable human demonstrations of coordinated articulated-object manipulation. Building on such references, physics-based methods learn coordinated contact behaviors through simulated hand--object interaction. ArtiGrasp~\cite{zhang2024artigrasp} uses reinforcement learning in physics simulation to synthesize bimanual grasping and articulation from a reference hand pose. For retargeting complete demonstration sequences, DexMachina~\cite{zhao2026dexmachina} combines motion and contact rewards with virtual object assistance, enabling policies to practice the reference motions before taking over the manipulation. This eases exploration, but assistance applied directly to the object leaves the required robot contact forces implicit. Wrench-based contact rewards account for the forces and torques that contacts can support, allowing human and robot interactions to be compared despite different contact geometries~\cite{zhu2026chord}. Such rewards evaluate contact capability; they do not estimate compensatory forces for correcting the current motion deviation. FoLD complements this supervision by estimating a compensatory force field over the object surface and providing it directly to the policy for corrective action generation.

\section{Method}
\label{sec:method}

\begin{figure*}[!t]
    \centering
    \includegraphics[width=\linewidth]{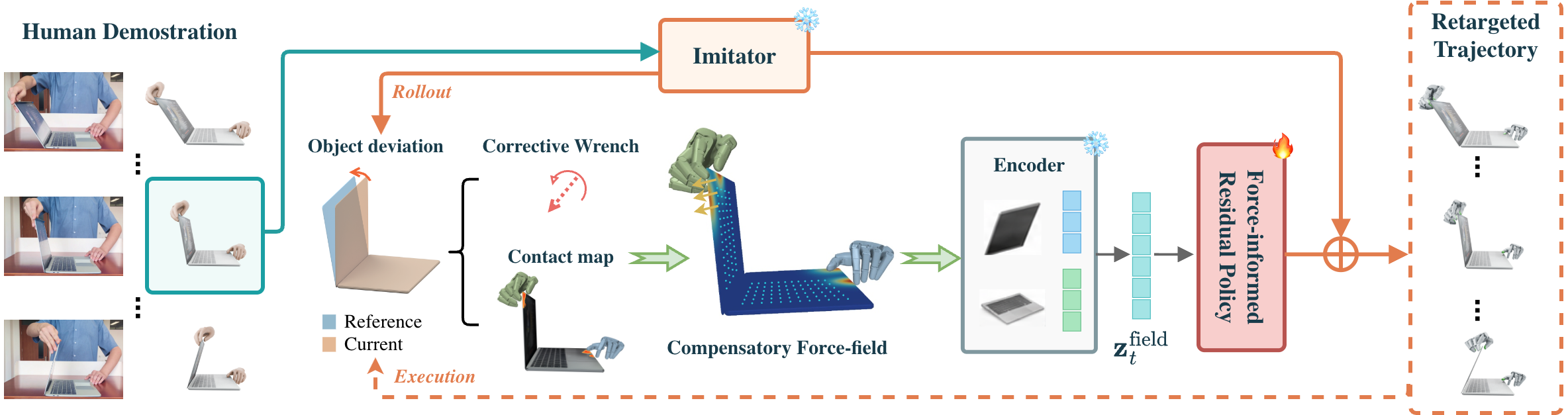}
    \vspace{-2em}
    \caption{Overview of FoLD. The imitator provides reference robot actions from a human demonstration. As the robot executes these actions, FoLD estimates a compensatory force field from the resulting object-motion deviations and the current contact map. The field is encoded into a latent representation that informs a residual policy, which corrects the reference actions to bring the object motion closer to the demonstration.}
    \label{fig:method_overview}
\end{figure*}

\subsection{Problem Formulation}
\label{sec:problem_formulation}

Given robot and object models and a bimanual human demonstration, we aim to train a policy $\pi_{\theta}$ that retargets human hand motions to robot hands while preserving the demonstrated interaction, where $\theta$ denotes the policy parameters. At execution step $t$, the robot action is sampled as $\mathbf{a}_t\sim\pi_{\theta}(\cdot\mid\mathbf{s}_t)$, where $\mathbf{s}_t$ denotes the environment state, including the object and hand states. For notational convenience, superscript $*$ denotes demonstrated quantities and $j$ indexes reference frames. We write the demonstration as
\begin{equation}
    \tau^{\mathrm{ref}}
    = \{(\mathbf{h}^{*}_{j},\mathbf{q}^{*}_{j})\}_{j=0}^{T-1},
\label{eq:reference_sequence}
\end{equation}
where $T$ is the sequence length, and $\mathbf{h}^{*}_{j}$ and $\mathbf{q}^{*}_{j}$ denote the bimanual hand pose and object configuration.
For both reference and executed motion, we define the object configuration as follows:
\begin{equation}
    \mathbf{q}_t=(\mathbf{p}_t,\mathbf{R}_t,\boldsymbol{\alpha}_t)
    \in\mathbb{R}^{3}\times\mathrm{SO}(3)\times\mathbb{R}^{d},
\label{eq:object_configuration}
\end{equation}
where $\mathbf{p}_t$ and $\mathbf{R}_t$ are the world-frame root position and orientation, and $\boldsymbol{\alpha}_t$ contains the articulation coordinates ($d=1$ for the hinged objects considered here).

\subsection{Overview}
\label{sec:method_overview}

Retargeting the demonstrated hand motion does not by itself ensure that the object follows the desired trajectory. Our key insight is to guide corrections to the hand actions with a compensatory force field that describes where and how forces should be applied to recover the demonstrated object motion.

Inspired by ManipTrans~\cite{li2025maniptrans}, we first train a hand imitator for each demonstration to retarget the hand motion while penalizing hand--object penetration. During this training, the object is prescribed to follow its demonstrated trajectory. However, at rollout time, executing the hand actions produced by the imitator may cause the object state at the next time step to deviate from the corresponding reference state in the demonstration.

To compensate for this deviation, we compute a corrective wrench from the difference between the current and reference object states. We then estimate a compensatory force field whose combined force and torque approximate this wrench, restricting the force distribution to regions indicated by a contact map computed from the current robot hand state. The field is encoded into a latent representation that guides the force-informed residual policy in predicting corrections to the imitator's reference actions, with the aim of recovering the demonstrated object motion.
The overall pipeline is illustrated in ~\autoref{fig:method_overview}. We describe compensatory force-field estimation in Sec.~\ref{sec:method_force_estimation}, representation learning in Sec.~\ref{sec:method_field_representation}, and learning of the force-informed residual policy in Sec.~\ref{sec:method_residual_learning}.

\subsection{Compensatory Force-Field Estimation}
\label{sec:method_force_estimation}

Information about the spatial distribution of contact forces can help a policy learn how to interact with an object to accomplish a task. For example, when opening a box, one hand supports the base while the other applies force to the lid; the distribution of these forces affects both object stability and task success. Human demonstrations, however, provide only kinematic information. We therefore explore whether a suitable compensatory force field can be inferred from this information to drive the object along the desired trajectory. Specifically, we compute a target wrench from object-motion deviations and a contact map from the current robot hand state. The wrench specifies the overall corrective force and torque, while the contact map constrains the regions of the object surface over which forces can be allocated. We then solve a regularized force-allocation problem within these regions under normal-force and friction constraints.

\subsubsection{Corrective Wrench}
Given a frozen imitator, we roll out a reference action at each execution step and compute a corrective wrench to compensate for the resulting deviation of the object state from the demonstration. We denote the object root state by $\mathbf{x}_t=(\mathbf{p}_t,\mathbf{R}_t,\mathbf{v}_t,\boldsymbol{\omega}_t)$, where $\mathbf{v}_t$ and $\boldsymbol{\omega}_t$ denote root linear and angular velocities. We index the reference by episode progress and use the next frame in the demonstration sequence as the target $(\mathbf{p}^{\mathrm{tar}}_t,\mathbf{R}^{\mathrm{tar}}_t)$, with reference velocities $(\mathbf{v}^{\mathrm{tar}}_t,\boldsymbol{\omega}^{\mathrm{tar}}_t)$ obtained by finite differences over the control interval. We construct the command in the world frame as
\begin{equation}
\begin{aligned}
    \mathbf{F}^{\mathrm{cmd}}_t
    &= k_p(\mathbf{p}^{\mathrm{tar}}_t-\mathbf{p}_t)
       + k_d(\mathbf{v}^{\mathrm{tar}}_t-\mathbf{v}_t),\\
    \boldsymbol{\tau}^{\mathrm{cmd}}_t
    &= k_R\operatorname{Log}\!\left(
        \mathbf{R}^{\mathrm{tar}}_t\mathbf{R}_t^{\top}\right)^{\vee}
       + k_{\omega}(\boldsymbol{\omega}^{\mathrm{tar}}_t
                     -\boldsymbol{\omega}_t),
\end{aligned}
\label{eq:field_feedback}
\end{equation}
where $\operatorname{Log}(\cdot)^{\vee}$ returns a rotation vector and the gains set the feedback strength. The pose terms drive the object toward the reference, while the velocity terms account for how it is already moving. We transform the corrective force and torque from the world frame into the current object frame and stack them into a wrench:
\begin{equation}
    \widetilde{\mathbf{w}}_t
    = \begin{bmatrix}
        \mathbf{R}_t^{\top}\mathbf{F}^{\mathrm{cmd}}_t\\
        \mathbf{R}_t^{\top}\boldsymbol{\tau}^{\mathrm{cmd}}_t
      \end{bmatrix}\in\mathbb{R}^{6}.
\label{eq:wrench_frame_transform}
\end{equation}
The first three components represent force, and the last three represent torque. We then define $\mathbf{w}^{\mathrm{cmd}}_t\in\mathbb{R}^{6}$ as the corrective wrench supplied to force allocation after limiting each component:
\begin{equation}
    w^{\mathrm{cmd}}_{t,i}
    = \min\!\left(w_i^{\max},
      \max\!\left(-w_i^{\max},\widetilde{w}_{t,i}\right)\right),
\label{eq:wrench_component_limits}
\end{equation}
where $i\in\{1,\ldots,6\}$ and $w_i^{\max}>0$ is the magnitude limit for the corresponding force or torque component. Components within $[-w_i^{\max},w_i^{\max}]$ remain unchanged, while those outside are set to the nearest bound. Recomputing this wrench at each control step makes the subsequent force allocation responsive to execution deviations.

\subsubsection{Anchor-Based Force Allocation}
Given the corrective wrench $\mathbf{w}^{\mathrm{cmd}}_t$ and contact map $\mathcal H_t$, we seek force-field coefficients whose induced forces reproduce the corrective wrench while remaining within the permitted contact regions. We formulate this allocation as a constrained optimization problem that balances wrench matching and force magnitude.

We represent the field at $N$ anchors on the object surface. Anchor $k$ has position $\mathbf c_{t,k}$, inward unit normal $\mathbf n_{t,k}$, and part label $b_k$. Positions and normals are expressed in the object frame. The contact map defines a surface region $\mathcal R_t$ and a binary mask
\begin{equation}
    \chi_{t,k}=\mathbf{1}[\mathbf c_{t,k}\in\mathcal R_t],
    \qquad k=1,\ldots,N,
\label{eq:anchor_contact_mask}
\end{equation}
which specifies whether anchor $k$ is available for force allocation.

The optimization variable is the coefficient vector $\boldsymbol\xi_t\in\mathbb R^{d_\xi}$, where $d_\xi$ is the number of field coefficients. A basis matrix $\mathbf B_t\in\mathbb R^{3N\times d_\xi}$ maps these coefficients to the stacked anchor forces:
\begin{equation}
    \mathbf f_t=\mathbf B_t\boldsymbol\xi_t,
    \qquad
    \mathbf f_{t,k}=\mathbf B_{t,k}\boldsymbol\xi_t,
\label{eq:field_coefficient_map}
\end{equation}
where $\mathbf B_{t,k}$ contains the three rows corresponding to anchor $k$. The resultant wrench is
\begin{equation}
    \mathbf G_t\mathbf B_t\boldsymbol\xi_t
    =\sum_{k=1}^{N}
    \begin{bmatrix}
        \mathbf f_{t,k}\\
        \mathbf c_{t,k}\times\mathbf f_{t,k}
    \end{bmatrix},
\label{eq:anchor_wrench_map}
\end{equation}
where $\mathbf G_t\in\mathbb R^{6\times3N}$ maps the anchor forces to total force and torque, with torque measured about the origin of the anchor coordinate frame.

We solve for the coefficients through
\begin{equation}
\begin{aligned}
    \hat{\boldsymbol\xi}_t
    =\arg\min_{\boldsymbol\xi_t}\quad
      &\frac{1}{2}\left\|\mathbf G_t\mathbf B_t\boldsymbol\xi_t
             -\mathbf w_t^{\mathrm{cmd}}\right\|_2^2+\frac{\lambda}{2}\left\|\mathbf B_t\boldsymbol\xi_t\right\|_2^2\\
    \mathrm{s.t.}\quad
      &\mathbf B_{t,k}\boldsymbol\xi_t=\mathbf 0, \quad
        \text{if }\chi_{t,k}=0,\\
      &\mathbf B_{t,k}\boldsymbol\xi_t\in\mathcal P_{t,k}, \quad
        \text{if }\chi_{t,k}=1.
\end{aligned}
\label{eq:force_allocation}
\end{equation}
The first term fits the combined force and torque to the corrective wrench. Because multiple force distributions can explain the same wrench, the second term penalizes large forces, with $\lambda>0$ controlling this trade-off. The first constraint fixes forces outside the contact region to zero. For the remaining anchors, $\mathcal P_{t,k}$ is the polyhedral set of admissible forces: it enforces the normal-force bound $0\leq\mathbf n_{t,k}^{\top}\mathbf f_{t,k}\leq f_{\max}$ and linear friction inequalities that limit tangential loading according to the friction coefficient $\mu$.

With a quadratic objective and linear constraints, \autoref{eq:force_allocation} is a convex quadratic program. As summarized in~\autoref{alg:force_field}, we use OSQP~\cite{stellato2020osqp} to solve for $\hat{\boldsymbol\xi}_t$ and recover the compensatory field as $\hat{\mathbf f}_t=\mathbf B_t\hat{\boldsymbol\xi}_t$. 
For the matrix construction in~\autoref{alg:force_field}, we express each admissible-force set as $\mathcal P_{t,k}=\{\mathbf f\mid\mathbf l^c_{t,k}\leq\mathbf C_{t,k}\mathbf f\leq\mathbf u^c_{t,k}\}$. The assembled QP minimizes $\tfrac12\boldsymbol\xi^{\top}\mathbf Q_t\boldsymbol\xi+\mathbf g_t^{\top}\boldsymbol\xi$ subject to $\mathbf l_t\leq\mathbf A_t\boldsymbol\xi\leq\mathbf u_t$. We use $[\mathbf c]_\times\mathbf v=\mathbf c\times\mathbf v$ and denote vertical block concatenation by $\operatorname{vstack}$.
The complete force-field computation takes approximately 60--120\,ms per frame, with an overall update rate of approximately 10--11\,Hz in our implementation. This allows the field to be updated online during policy training, providing spatial force guidance that adapts to the current hand-object interaction.

\begin{algorithm}[h]
\caption{Compensatory Force-Field Allocation}
\label{alg:force_field}
\small
\begin{algorithmic}[1]
\REQUIRE Contact map $\mathcal H_t$, wrench $\mathbf w_t^{\mathrm{cmd}}$, basis $\mathbf B_t$, $\lambda>0$; anchor data $\{\mathbf c_{t,k},\mathbf C_{t,k},\mathbf l^c_{t,k},\mathbf u^c_{t,k}\}_{k=1}^{N}$
\ENSURE Compensatory force field $\hat{\mathbf f}_t$
\STATE $\mathcal R_t\gets\operatorname{ContactRegion}(\mathcal H_t)$
\FOR{$k=1,\ldots,N$}
    \STATE $\chi_{t,k}\gets\mathbf 1[\mathbf c_{t,k}\in\mathcal R_t]$
    \STATE $\mathbf G_{t,k}\gets\left[\begin{smallmatrix}\mathbf I_3\\{[\mathbf c_{t,k}]_\times}\end{smallmatrix}\right]$
    \IF{$\chi_{t,k}=0$}
        \STATE $\mathbf A_{t,k}\gets\mathbf B_{t,k},\quad\mathbf l_{t,k}\gets\mathbf u_{t,k}\gets\mathbf 0_3$
    \ELSE
        \STATE $\mathbf A_{t,k}\gets\mathbf C_{t,k}\mathbf B_{t,k}$
        \STATE $(\mathbf l_{t,k},\mathbf u_{t,k})\gets(\mathbf l^c_{t,k},\mathbf u^c_{t,k})$
    \ENDIF
\ENDFOR
\STATE $\mathbf G_t\gets[\mathbf G_{t,1}\ \cdots\ \mathbf G_{t,N}]$
\STATE $\mathbf A_t\gets\operatorname{vstack}_{k=1}^{N}\mathbf A_{t,k}$
\STATE $\mathbf l_t\gets\operatorname{vstack}_{k=1}^{N}\mathbf l_{t,k},\quad\mathbf u_t\gets\operatorname{vstack}_{k=1}^{N}\mathbf u_{t,k}$
\STATE $\mathbf Q_t\gets\mathbf B_t^{\top}(\mathbf G_t^{\top}\mathbf G_t+\lambda\mathbf I_{3N})\mathbf B_t$
\STATE $\mathbf g_t\gets-\mathbf B_t^{\top}\mathbf G_t^{\top}\mathbf w_t^{\mathrm{cmd}}$
\STATE $\hat{\boldsymbol\xi}_t\gets\operatorname{OSQP}(\mathbf Q_t,\mathbf g_t,\mathbf A_t,\mathbf l_t,\mathbf u_t)$
\STATE $\hat{\mathbf f}_t\gets\mathbf B_t\hat{\boldsymbol\xi}_t$
\RETURN $\hat{\mathbf f}_t$
\end{algorithmic}
\end{algorithm}

\begin{figure*}[t]
    \centering

    \begingroup
    \scriptsize

    \noindent
    \makebox[0.25\textwidth][c]{Human}%
    \makebox[0.25\textwidth][c]{DexMachina}%
    \makebox[0.25\textwidth][c]{Hand-only Imitation}%
    \makebox[0.25\textwidth][c]{Ours}%

    \endgroup

    \vspace{1mm}

    \includegraphics[width=\textwidth]{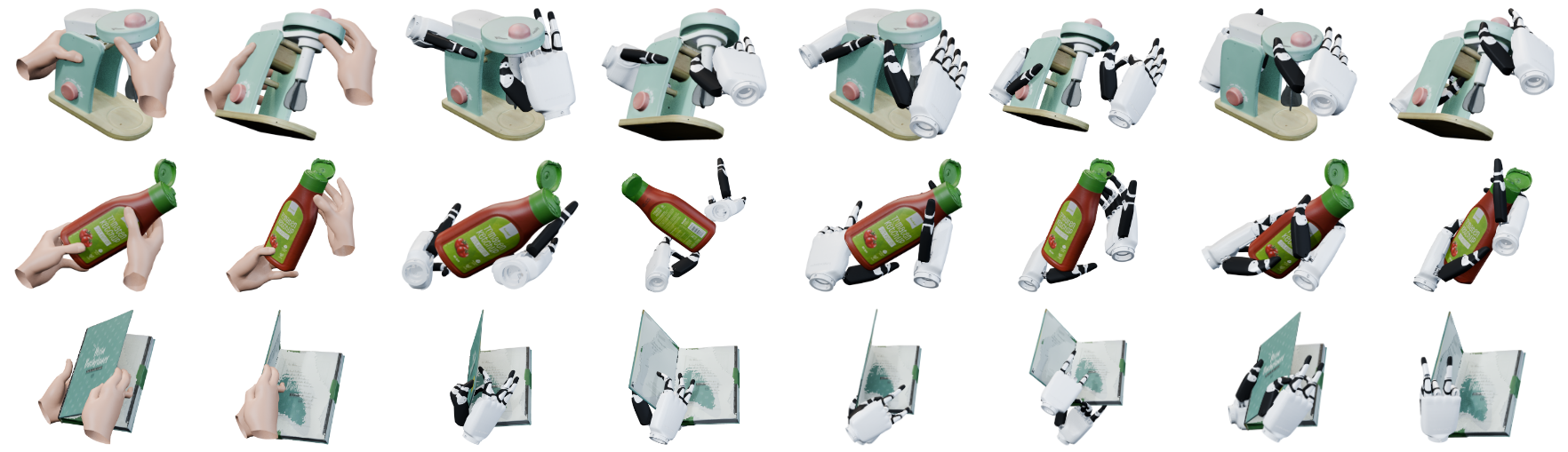}
    \vspace{-2em}
    \caption{Qualitative results on mixer, ketchup-bottle, and notebook manipulation (top to bottom). From left to right: human demonstrations, DexMachina, hand-only imitation, and FoLD, with two snapshots per group. Hand-only imitation uses only our pretrained imitator's hand actions, with the object's root pose and articulation fixed to the demonstrated state at each frame.}
    \label{fig:qualitative_results}
\end{figure*}

\subsection{Compensatory Force-Field Representation Learning}
\label{sec:method_field_representation}

The estimated compensatory force field contains a three-dimensional force at each surface anchor, resulting in a high-dimensional input for the force-informed residual policy. We therefore use an encoder--decoder model to learn a low-dimensional representation of the field through reconstruction.

For articulated objects, forces applied to different parts can play distinct roles in the task. We therefore use a PointNet-style encoder~\cite{qi2017pointnet} that combines global and part-level features to capture both the overall force distribution and how forces are distributed across individual parts, producing a compact latent representation:
\begin{equation}
    \mathbf{z}^{\mathrm{field}}_t
    = E_{\phi}\!\left(
        \{\mathbf{c}_{t,k},\mathbf{n}_{t,k},
          \hat{\mathbf{f}}_{t,k},b_k,m_{t,k}\}_{k=1}^{N}
      \right),
\end{equation}
where $\phi$ denotes the encoder parameters and $m_{t,k}$ indicates whether the force magnitude exceeds a small threshold. This encoding provides a fixed-dimensional policy input that is invariant to anchor ordering.

We train the encoder--decoder using compensatory force fields generated offline from the kinematic demonstrations. Each sample provides anchor forces and their corresponding generalized force as reconstruction targets.
A local decoder reconstructs each anchor force from the latent vector and anchor geometry, encouraging the encoding to preserve the spatial force distribution. To also supervise the combined mechanical effect, an auxiliary head predicts the corresponding generalized force. For the hinged objects considered here, this seven-dimensional quantity comprises the root force and torque together with the hinge torque. Both branches are used only during representation training. We optimize
\begin{equation}
    \mathcal{L}_{\mathrm{enc}}
    =\mathcal{L}_{\mathrm{force}}
     +\beta_w\mathcal{L}_{\mathrm{wrench}}
     +\beta_d\mathcal{L}_{\mathrm{direction}},
\label{eq:field_encoder_loss}
\end{equation}
where $\mathcal{L}_{\mathrm{force}}$ and $\mathcal{L}_{\mathrm{wrench}}$ measure the reconstruction errors of the anchor forces and the generalized force, respectively, using SmoothL1 loss. The cosine loss $\mathcal{L}_{\mathrm{direction}}$ additionally penalizes differences between the predicted and target force directions. By training this encoder--decoder, we obtain a compact latent representation of the compensatory force field. Appendix~\ref{app:field_encoder} provides the feature aggregation and pretraining details.

\subsection{Force-informed Residual Policy}
\label{sec:method_residual_learning}
The compensatory force field indicates how contact forces could compensate for deviations from the demonstrated object motion. To translate this guidance into robot actions, we condition a force-informed residual policy on the field representation $\mathbf{z}^{\mathrm{field}}_t$, alongside the current state and demonstration reference. The policy predicts corrections that are added to the frozen imitator's reference actions, allowing force information to guide adjustments to the hand--object interaction. The residual is scaled by $s=0.1$ and initialized to produce zero corrections; Appendix~\ref{app:residual_policy} gives the action composition and architecture details.

We train the residual policy using proximal policy optimization (PPO)~\cite{schulman2017proximal}. The reward combines object-root position, orientation, and articulation tracking with hand-pose imitation, retargeted joint-reference tracking, and position-based hand--object contact tracking, following DexMachina~\cite{zhao2026dexmachina}. FoLD additionally rewards frame-to-frame consistency of wrist and fingertip motion with the demonstration. These terms encourage task completion while preserving the demonstrated manipulation pattern, and contact-force and action-magnitude penalties discourage excessive contact forces and large control inputs. Reward definitions, coefficients, and tracking scales are given in Appendix~\ref{app:reward_config} and \autoref{tab:reward_weights}.

During residual-policy training, we also adopt the virtual object controller (VOC) curriculum from DexMachina~\cite{zhao2026dexmachina}. The VOC initially assists the object in following the demonstrated motion, with assistance progressively reduced to zero. After VOC assistance reaches zero, we continue training the residual policy with the VOC disabled. The VOC is also disabled throughout evaluation, so object motion is produced entirely through the robot's hand--object interaction. Appendix~\ref{app:voc} details the curriculum schedule.

\section{Experiments}
\label{sec:result}
Our experiments evaluate FoLD's ability to reproduce human-demonstrated articulated-object manipulation across robotic hands. Comparisons with DexMachina establish the overall gains in task completion and motion tracking; integration with CHORD and controlled ablations examine the additional value of compensatory force-field input. Physical robot demonstrations further evaluate whether the generated motions preserve the coordination needed to execute the demonstrated tasks on hardware.

\subsection{Experimental Setup}
\label{sec:exp_setup}

\textbf{Tasks and embodiments.}
Our main simulation benchmark uses seven clips from ARCTIC~\cite{fan2023arctic}, covering a box, a ketchup bottle, a mixer, a notebook, and a waffle iron. The clips contain 100--300 reference frames and include both object-root motion and articulation. We compare FoLD with DexMachina~\cite{zhao2026dexmachina} on Inspire, Allegro, XHand, and Schunk hands, training a separate policy for each hand--task--method configuration. Within each comparison, the methods share demonstrations, retargeted references, initial states, and physical parameters.
The clip definitions are provided in Appendix~\ref{app:dex_tasks}.
We additionally evaluate force-informed learning within CHORD~\cite{zhu2026chord} using the Sharpa Wave hand, as detailed in Sec.~\ref{sec:exp_chord}.

\textbf{Training and evaluation.}
The main experiments use Genesis~\cite{genesis2024} and PPO~\cite{schulman2017proximal} with 4,096 parallel environments at $60\,\mathrm{Hz}$. FoLD freezes its pretrained hand-motion imitator and force-field encoder during residual-policy training. DexMachina and FoLD use a virtual object controller (VOC) curriculum and matched policy-training budgets. We select the checkpoint with the highest recorded mean training reward among those saved after VOC assistance reaches zero. Quantitative evaluation uses deterministic actions with environment randomization and the VOC disabled. Thus, the reported object motion is produced entirely through the robot's hand--object interaction; the compensatory field supplies information to the policy without applying forces to the object. Appendix~\ref{app:training} gives the training stages, budgets, rewards, and curriculum settings.

\textbf{Metrics.}
We measure task completion, object-motion accuracy, and contact quality. Following CHORD~\cite{zhu2026chord}, success rate (SR) is the completion length ratio $L/T$: $L$ counts consecutive frames before root-position error exceeds $15\,\mathrm{cm}$ or root-rotation error exceeds $40^\circ$, and $T$ is the full evaluation length. Mean root-position (cm), root-rotation (degrees), and articulation (Joint, degrees) errors quantify motion accuracy over the full trajectory, including frames after a threshold violation. Contact Mean measures agreement with demonstrated contact locations in the DexMachina comparison. For CHORD, Contact Wrench Support Mean (CWS Mean) measures how well contacts support the force and torque demands of the demonstrated motion. Higher contact scores indicate better agreement or support within the respective comparison. Per-hand results are macro-averages over tasks; overall results weight all included hand--task pairs equally. Appendix~\ref{app:metrics} provides the full definitions.

\subsection{Comparison across Hand Embodiments}
\label{sec:exp_dexmachina}

FoLD achieves higher overall task completion and more accurate object motion than DexMachina (\autoref{tab:comparison_dexmachina}). Overall SR increases from 0.606 to 0.772, accompanied by lower root-position, root-rotation, and articulation errors. The 52.3\% reduction in articulation error is particularly relevant for these tasks: reproducing the demonstrated interaction requires controlling the relative motion of object parts as well as the object's global pose. Higher Contact Mean further indicates closer agreement with the demonstrated contact pattern. Taken together, these results show that the gain in completion is accompanied by better preservation of the intended manipulation.

\begin{table}[t]
\centering
\caption{Comparison of DexMachina and FoLD across robotic hands. Both methods use the same 27 paired hand--task evaluations (aggregation details in Appendix~\ref{app:divergent_case}). Best results for each hand are shown in bold.}
\label{tab:comparison_dexmachina}

\renewcommand{\arraystretch}{1.25}
\resizebox{0.9\columnwidth}{!}{
\begin{tabular}{ccccccc}
\toprule
\textbf{Hand} & \textbf{Method}
& \textbf{SR} $\uparrow$
& \begin{tabular}[c]{@{}c@{}}\textbf{Contact}\\\textbf{Mean} $\uparrow$\end{tabular}
& \begin{tabular}[c]{@{}c@{}}\textbf{Pos.}\\\textbf{(cm)} $\downarrow$\end{tabular}
& \begin{tabular}[c]{@{}c@{}}\textbf{Rot.}\\\textbf{($^\circ$)} $\downarrow$\end{tabular}
& \begin{tabular}[c]{@{}c@{}}\textbf{Joint}\\\textbf{($^\circ$)} $\downarrow$\end{tabular} \\
\midrule

\multirow{2}{*}{Inspire}
& DexMachina
& 0.511 & 0.154 & 52.27 & 42.65 & 34.26 \\

& \textbf{Ours}
& \textbf{0.700}
& \textbf{0.248}
& \textbf{18.19}
& \textbf{23.28}
& \textbf{20.20} \\

\hdashline[0.5pt/5pt]

\multirow{2}{*}{Allegro}
& DexMachina
& 0.596
& 0.043
& \textbf{20.07}
& 49.38
& 22.51 \\

& \textbf{Ours}
& \textbf{0.750}
& \textbf{0.061}
& 23.62
& \textbf{40.51}
& \textbf{12.18} \\

\hdashline[0.5pt/5pt]

\multirow{2}{*}{XHand}
& DexMachina
& \textbf{0.724}
& 0.113
& 39.55
& 48.34
& 24.88 \\

& \textbf{Ours}
& 0.697
& \textbf{0.116}
& \textbf{31.86}
& \textbf{43.76}
& \textbf{20.24} \\

\hdashline[0.5pt/5pt]

\multirow{2}{*}{Schunk}
& DexMachina
& 0.588
& 0.034
& 34.47
& 45.66
& 44.49 \\

& \textbf{Ours}
& \textbf{0.968}
& \textbf{0.101}
& \textbf{0.91}
& \textbf{6.52}
& \textbf{5.37} \\

\hline

\multirow{2}{*}{Overall}
& DexMachina
& 0.606
& 0.088
& 36.67
& 46.54
& 31.06 \\

& \textbf{Ours}
& \textbf{0.772}
& \textbf{0.133}
& \textbf{19.30}
& \textbf{29.33}
& \textbf{14.83} \\

\bottomrule
\end{tabular}
}
\end{table}

FoLD's advantage over DexMachina extends across the four evaluated hand embodiments. Contact consistency, object-root rotation accuracy, and articulation accuracy improve on every hand, showing that the method more faithfully reproduces the demonstrated interaction across different hand designs. This consistency supports the use of object-surface force guidance for cross-embodiment manipulation: the field describes the desired mechanical correction of the object, while each hand-specific policy learns how to realize it through its own finger motions and contacts. FoLD therefore provides an effective way to adapt the same demonstrated manipulation to multiple robotic hand embodiments.

We present qualitative results in \autoref{fig:qualitative_results} on mixer, ketchup-bottle, and notebook manipulation, comparing DexMachina, hand-only imitation, and FoLD with the human demonstrations. \emph{Hand-only imitation} denotes the execution of hand actions generated solely by our pretrained hand-motion imitator, without residual corrections. In this reference, the object's root pose and articulation are fixed to their demonstrated values at each frame, visualizing the learned hand-motion prior around a prescribed object trajectory. DexMachina and FoLD manipulate a dynamic object through hand contacts with the VOC disabled.

\begin{figure}[t]
    \centering

    \begingroup
    \scriptsize

    \noindent
    \makebox[0.2\linewidth][c]{Human}%
    \makebox[0.2\linewidth][c]{Allegro Hand}%
    \makebox[0.2\linewidth][c]{Inspire Hand}%
    \makebox[0.2\linewidth][c]{Schunk Hand}%
    \makebox[0.2\linewidth][c]{XHand}%

    \endgroup

    \vspace{1mm}

    \includegraphics[width=\linewidth]{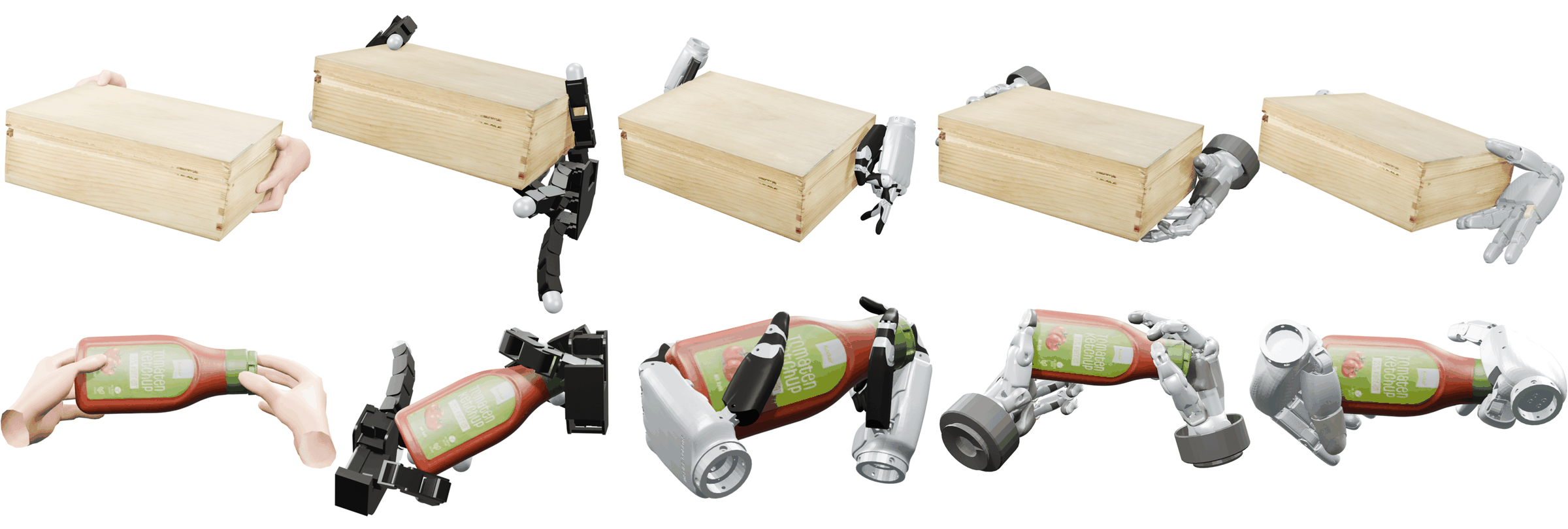}
    \vspace{-2em}
    \caption{Bimanual manipulation across hand embodiments.
    Human configurations appear on the left, followed by robot configurations learned with FoLD.}
    \label{fig:cross_hand}
\end{figure}

\begin{figure}[t]
    \centering
    \includegraphics[width=\linewidth]{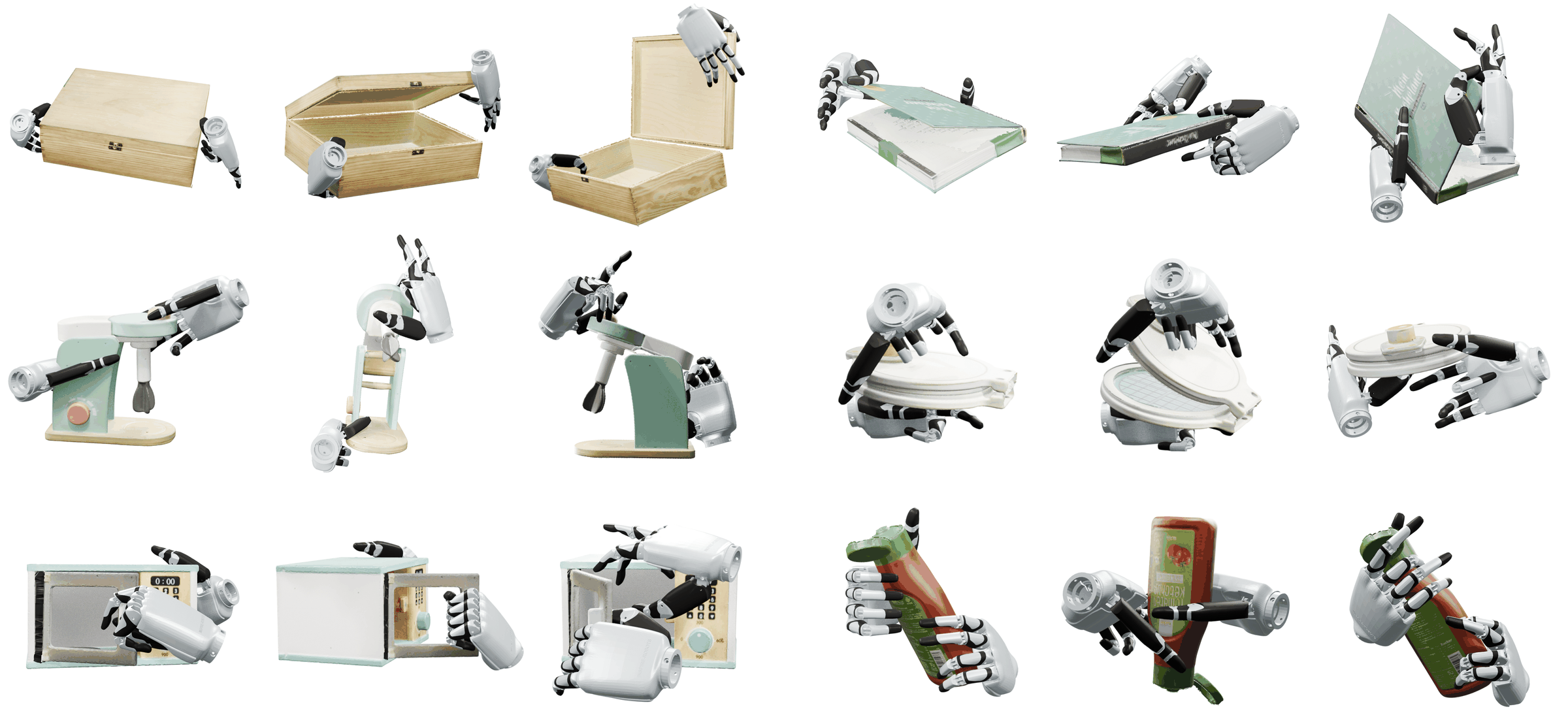}
    \caption{Additional FoLD manipulation examples. Each group presents three snapshots showing coordinated support and manipulation of articulated object parts.}
    \label{fig:gallery}
\end{figure}

FoLD reproduces the demonstrated division of roles between the two hands using robot-specific finger configurations. In the notebook example, one hand supports one cover while the other manipulates the hinged cover, retaining the intended interaction between the object parts. The different finger placements illustrate how FoLD adapts the demonstration to the robot while preserving the manipulation goal. \autoref{fig:cross_hand} shows this coordination realized by different hand embodiments, and \autoref{fig:gallery} extends the qualitative examples to additional articulated objects. These results provide visual evidence that FoLD preserves the functional hand--object relationships underlying the demonstrated manipulation.

\subsection{Force-Informed Learning within CHORD}
\label{sec:exp_chord}

We further evaluate whether our force-field representation can improve an existing contact-guided method. We augment CHORD~\cite{zhu2026chord} with our compensatory force-field latent as an additional policy input, forming \textbf{CHORD + FoLD}. This integration retains CHORD's native policy structure, rewards, and training protocol, allowing us to assess the benefit of the added force information. We compare both variants on five demonstration sequences with the Sharpa Wave hand; implementation and evaluation details are provided in Appendix~\ref{app:chord_details}.

\begin{figure}[t]
    \centering

    \begingroup
    \scriptsize

    \noindent
    \makebox[0.3333\linewidth][c]{Human}%
    \makebox[0.3333\linewidth][c]{\kern-3mm CHORD\kern3mm}%
    \makebox[0.3333\linewidth][c]{\kern-3mm CHORD + FoLD\kern3mm}%

    \endgroup

    \vspace{1mm}

    \includegraphics[width=\linewidth]{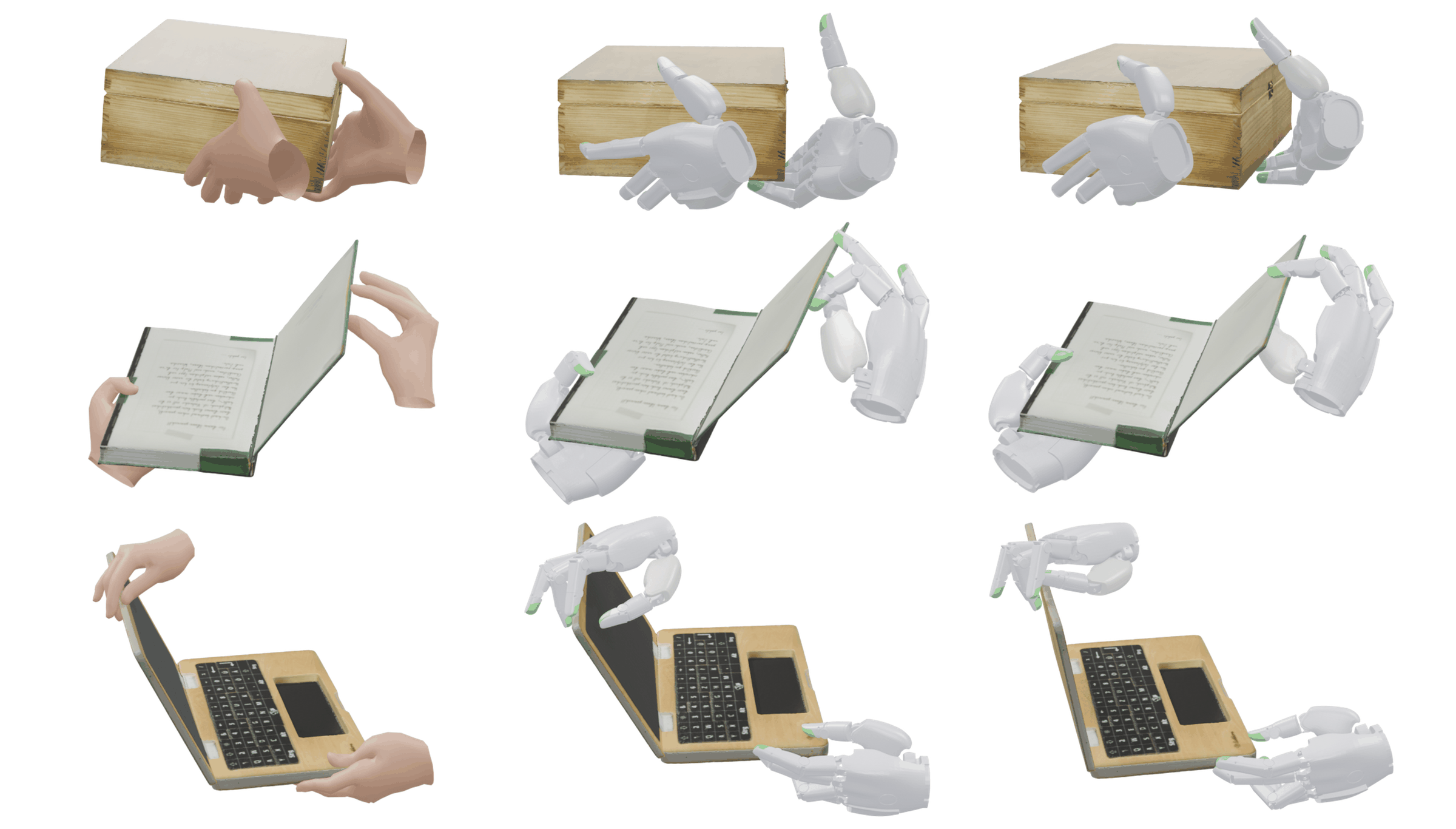}
    \vspace{-1.5em}
    \caption{Human demonstrations, CHORD, and CHORD + FoLD on box, notebook, and laptop manipulation (top to bottom). These snapshots illustrate hand configurations relative to the articulated object parts.}
    \label{fig:comparison_mano_chord_ours}
\end{figure}

\autoref{tab:comparison_chord} shows that CHORD + FoLD reduces mean object-root position and rotation errors by 38.6\% and 22.9\%, respectively, relative to CHORD. These gains demonstrate that our force-field input improves motion tracking even when the policy already learns from contact wrench-space guidance. The two forms of guidance serve complementary roles: CHORD's reward evaluates the mechanical support provided by the contacts, while our field informs the policy of the force adjustments that could correct the current object-motion deviation. Providing this corrective information enables more accurate reproduction of the demonstrated object motion under the same learning objective. \autoref{fig:comparison_mano_chord_ours} presents qualitative comparisons on box, notebook, and laptop manipulation. Together, these results support the value of our force-field representation as a policy input that can strengthen an existing contact-guided learning framework.

\begin{table}[t]
\caption{Comparison of CHORD and CHORD + FoLD on Sharpa-Wave dexterous hand. Values are mean $\pm$ sample standard deviation.}
\label{tab:comparison_chord}
\centering
\small
\renewcommand{\arraystretch}{1.2}
\setlength{\tabcolsep}{4pt}
\resizebox{\columnwidth}{!}{%
\begin{tabular}{lccccc}
\toprule
\textbf{Method}
& SR $\uparrow$
& \shortstack{CWS\\Mean $\uparrow$}
& \shortstack{Pos.\\(cm) $\downarrow$}
& \shortstack{Rot.\\($^\circ$) $\downarrow$}
& \shortstack{Joint\\($^\circ$) $\downarrow$} \\
\midrule
CHORD & $0.392{\pm}0.315$ & $\mathbf{0.351{\pm}0.242}$ & $32.90{\pm}29.47$ & $43.58{\pm}38.03$ & $\mathbf{21.92{\pm}23.45}$ \\
\textbf{CHORD + FoLD} & $\mathbf{0.397{\pm}0.314}$ & $0.344{\pm}0.225$ & $\mathbf{20.19{\pm}17.92}$ & $\mathbf{33.60{\pm}30.40}$ & $22.17{\pm}24.00$ \\
\bottomrule
\end{tabular}
}
\end{table}

\begin{figure*}
    \centering
    \includegraphics[width=\textwidth]{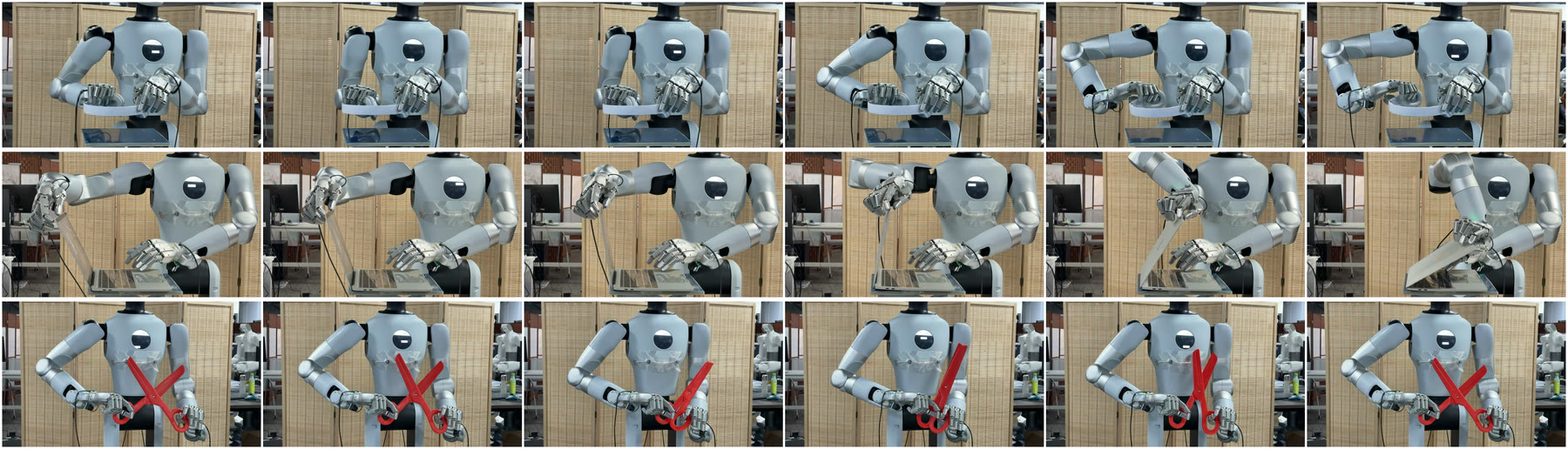}
    \caption{Real-robot deployment on a Galbot G3 equipped with two Galbot Sharpa hands. Each row shows six frames from one execution: headphone folding (top), laptop closing (middle), and scissors opening and closing (bottom).}
    \label{fig:real_robot_experiments_galbot}
\end{figure*}

\subsection{Ablation Study}
\label{sec:exp_ablation}

We evaluate the contribution of force-informed residual learning on Inspire using Box-200, Mixer-170, and Notebook-147. All three configurations train a residual policy on top of the same frozen imitator and share the residual architecture and scale, rewards, and training budget. Starting with neither VOC assistance nor force-field input, we first add the VOC curriculum adopted from DexMachina~\cite{zhao2026dexmachina}, then add the compensatory force-field input. The final configuration is FoLD as evaluated in the main comparison. Evaluation is deterministic with the VOC disabled. Appendix~\ref{app:ablation_tasks} gives the detailed configuration.

The VOC curriculum improves completion of the residual baseline, raising SR from 0.848 to 1.000 in \autoref{tab:ablation_pipeline}.

With the VOC curriculum held fixed, force-informed residual learning improves contact consistency and root-position accuracy. Adding the force-field input raises Contact Mean from 0.275 to 0.351 and reduces position error by 37.3\%, while retaining $\mathrm{SR}=1.000$. With the rewards and training setup held fixed, this comparison isolates the benefit of compensatory force-field conditioning: it supplies corrective information that helps the residual policy reproduce the demonstrated contact pattern and object position more accurately, beyond the completion already achieved with curriculum assistance.

\begin{table}[t]
\caption{Component ablation on Inspire. Checkmarks and crosses indicate enabled and disabled components, respectively. All rows use the same frozen imitator; the last row is FoLD as evaluated in the main comparison. Bold indicates the best result.}
\label{tab:ablation_pipeline}
\centering
\small
\renewcommand{\arraystretch}{1.2}
\setlength{\tabcolsep}{2.5pt}
\resizebox{\columnwidth}{!}{
\begin{tabular}{cccccccc}
\toprule
\shortstack{Residual\\policy}
& \shortstack{VOC\\curriculum}
& \shortstack{Force-field\\input}
& SR $\uparrow$
& \shortstack{Contact\\Mean $\uparrow$}
& \shortstack{Pos.\\(cm) $\downarrow$}
& \shortstack{Rot.\\($^\circ$) $\downarrow$}
& \shortstack{Joint\\($^\circ$) $\downarrow$} \\
\midrule
\cmark & \xmark & \xmark & 0.848 & 0.300 & 1.04 & 11.33 & \textbf{11.99} \\
\cmark & \cmark & \xmark & \textbf{1.000} & 0.275 & 1.34 & \textbf{5.55} & 13.30 \\
\cmark & \cmark & \cmark & \textbf{1.000} & \textbf{0.351} & \textbf{0.84} & 7.69 & 13.76 \\
\bottomrule
\end{tabular}
}
\end{table}

\subsection{Real-Robot Deployment}
\label{sec:exp_real}

We validate the physical executability of FoLD-generated manipulation motions on two hardware platforms: a Galbot G3 equipped with two Galbot Sharpa hands and a dual-arm setup comprising two Franka Research 3 manipulators with Sharpa Wave hands. Starting from in-house human videos reconstructed with ArtHOI~\cite{wang2026arthoi} and scanned object meshes, FoLD generates robot-hand trajectories in simulation. Inverse kinematics in Isaac Lab~\cite{mittal2025isaaclab} maps these hand targets to arm trajectories, which are executed together with the hand motions through open-loop replay. Appendix~\ref{app:real_robot} provides the deployment details.

\begin{figure}[t]
    \centering

    \begin{tikzpicture}
        \node[anchor=south west, inner sep=0] (img) at (0.09\linewidth,0) {
            \includegraphics[width=0.91\linewidth]{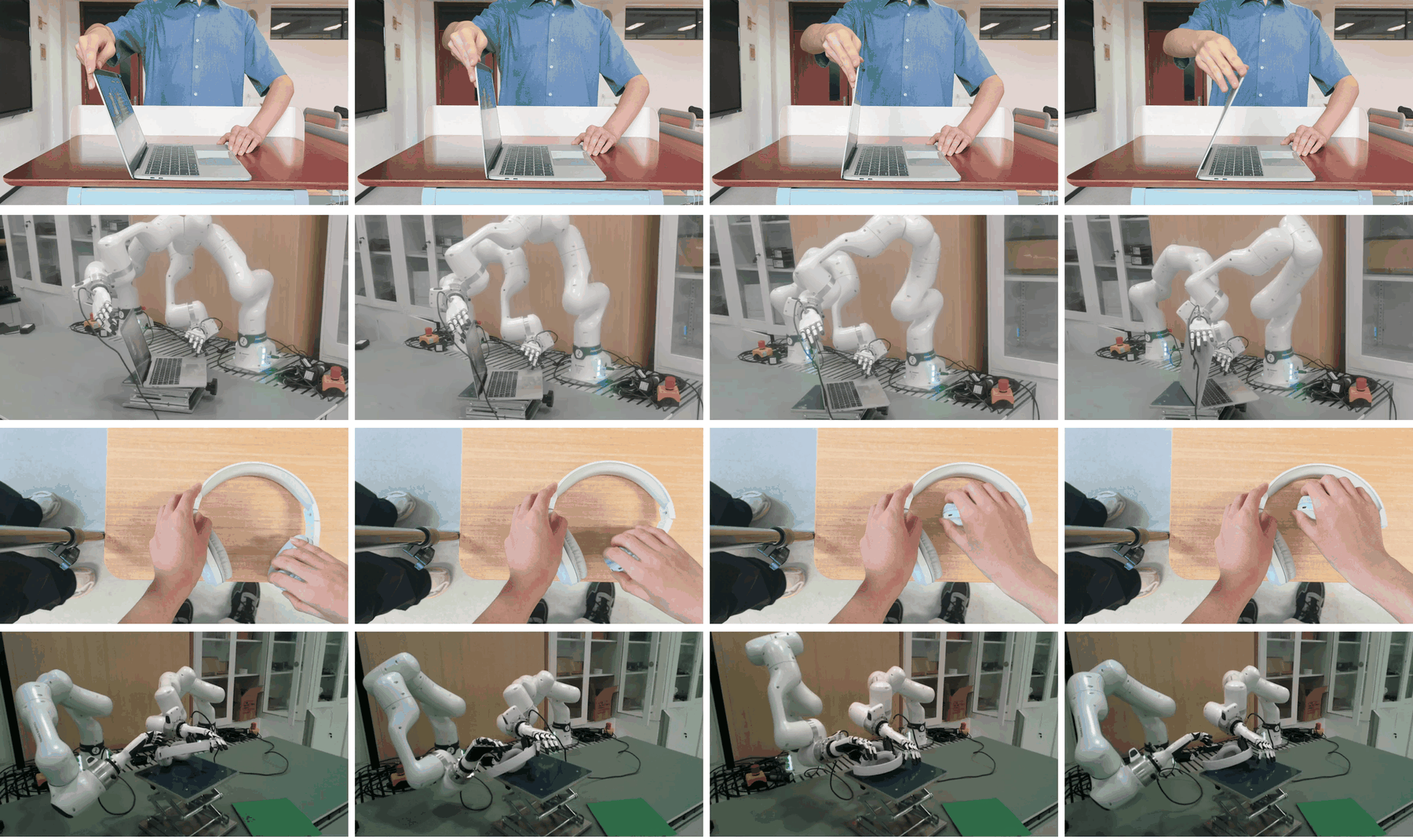}
        };
        \scriptsize

        \node[
            rotate=90,
            anchor=center,
            inner sep=0pt
        ] at
        ($ (img.south west)!0.875!(img.north west) + (-0.045\linewidth,0) $)
        {\shortstack{Human\\Demo}};

        \node[
            rotate=90,
            anchor=center,
            inner sep=0pt
        ] at
        ($ (img.south west)!0.625!(img.north west) + (-0.045\linewidth,0) $)
        {\shortstack{Robot\\Deployment}};

        \node[
            rotate=90,
            anchor=center,
            inner sep=0pt
        ] at
        ($ (img.south west)!0.375!(img.north west) + (-0.045\linewidth,0) $)
        {\shortstack{Human\\Demo}};

        \node[
            rotate=90,
            anchor=center,
            inner sep=0pt
        ] at
        ($ (img.south west)!0.125!(img.north west) + (-0.045\linewidth,0) $)
        {\shortstack{Robot\\Deployment}};

    \end{tikzpicture}
    \vspace{-1em}
    \caption{Real-robot deployment on two Franka Research 3 arms equipped with Sharpa Wave hands. The upper two rows show laptop closing, and the lower two rows show headphone folding; each task pairs the human demonstration with the corresponding robot execution.}
    \label{fig:real_robot_experiments_fr3}
\end{figure}

As shown in \autoref{fig:real_robot_experiments_galbot} and \autoref{fig:real_robot_experiments_fr3} , FoLD successfully folds headphones and closes a laptop on both platforms, and further opens and closes scissors on the Galbot G3; additional examples and failure cases are provided in the accompanying video.

\section{Conclusion}
\label{sec:conclusion}

We presented FoLD, a framework that uses explicit force guidance to retarget human demonstrations to dexterous robots for articulated-object manipulation. By conditioning a residual policy on compensatory force fields, FoLD connects the desired object motion to the mechanical adjustments needed during robot execution. Experiments across multiple hand embodiments show improved aggregate task completion and object tracking, while real-robot demonstrations show that the generated motions can be physically executed. Together, these results highlight the value of force-informed learning for adapting human manipulation skills to dexterous robots.

Despite these promising results, FoLD has several limitations. Our current formulation is state-based and relies on accurate reconstructions of human hand--object interactions. Each policy is also trained for a single object--task pair, limiting generalization. In future work, we will explore extending FoLD to visuomotor policy learning on real robots, aiming to improve robustness and generalization across objects and tasks.

\FloatBarrier
\bibliographystyle{IEEEtran}
\bibliography{main}

\clearpage
\useRomanappendicesfalse
\appendices

\section{Implementation Details}
\label{app:implementation}

\subsection{Online Compensatory Force-Field Construction}
\label{app:force_field}

At each compensatory force-field update, we compute a corrective object wrench from the
difference between the current object state and the next reference
state. The wrench is expressed in the object-aligned frame and
approximately distributed over a set of precomputed surface anchors.
The input contact map is converted to a binary mask over the anchors.
Anchors marked zero have their force fixed to zero and contribute
neither force nor torque to the wrench fit. We solve the resulting
masked coefficient problem using OSQP~\cite{stellato2020osqp}, as detailed in
\autoref{alg:force_field}. For articulated objects, the cached anchor geometry follows the
reference articulation at the corresponding episode index.

The resulting compensatory forces at the anchors are used only to construct
the policy conditioning input. They are not directly applied to the simulated
object and should not be interpreted as measured robot contact forces.
The online compensatory force-field parameters are summarized in
\autoref{tab:field_impl}.

\begin{table}[h]
    \centering
    \caption{Online compensatory force-field construction parameters.}
    \label{tab:field_impl}
    \begin{tabular}{lc}
        \toprule
        Parameter & Value \\
        \midrule
        Position proportional gain $k_p$       & 200 \\
        Position derivative gain $k_d$         & 20 \\
        Rotation proportional gain $k_R$       & 20 \\
        Rotation derivative gain $k_{\omega}$  & 2 \\
        Force command scale                    & $25\,\mathrm{N}$ \\
        Torque command scale                   & $3\,\mathrm{N\,m}$ \\
        Componentwise command limit            & 5 \\
        Friction coefficient $\mu$              & 0.5 \\
        Normal-force limit $f_{\max}$           & $100\,\mathrm{N}$ \\
        Regularization coefficient $\lambda$    & $10^{-4}$ \\
        \bottomrule
    \end{tabular}
\end{table}

\subsection{Compensatory Force-Field Encoding}
\label{app:field_encoder}

A separate compensatory force-field encoder is trained for each manipulation task.
Each encoder is pretrained for 300 epochs on offline compensatory force-field
solutions and then frozen throughout policy training. The encoder maps
the anchor-wise field, including anchor positions, surface normals,
allocated forces, activity indicators, and object-part identities, to
a 128-dimensional latent vector.

The encoder follows the shared per-point feature extraction and symmetric
aggregation of PointNet~\cite{qi2017pointnet}. Each anchor feature also includes
the log-transformed force magnitude and a learned object-part embedding.
We concatenate mean, max, and force-magnitude-weighted mean pooling
features computed globally and separately for each object part, then
map the pooled features to the latent vector. This aggregation preserves
both the overall force distribution and its allocation across articulated
parts while remaining invariant to anchor ordering.

For articulated objects, the offline generalized-force target includes
an additional articulation torque, whereas online field construction
uses a six-dimensional corrective root wrench. Both stages use the
same anchor-wise field representation. The compensatory force-field latent is
provided to the force-informed residual policy as an additional observation and does
not receive a separate compensatory force-field reward.

\subsection{Force-informed Residual Policy}
\label{app:residual_policy}

The hand-motion imitator and force-informed residual policy use separate feedforward
encoders for semantic observation fields. The imitator receives the
native robot--object observation, while the force-informed residual policy additionally
receives the 128-dimensional compensatory force-field latent. No separate recurrent
history encoder is used.

The final action mean is

\begin{equation}
    \boldsymbol{\mu}_t
    =
    \operatorname{clip}
    \left(
        \pi_{\mathrm{imit}}(\mathbf{o}_t)
        +
        s\,\pi_{\mathrm{res}}
        (\mathbf{o}_t,\mathbf{z}^{\mathrm{field}}_t),
        -1,1
    \right),
\end{equation}

where the residual scale is $s=0.1$. The imitator is frozen during
residual learning, and the output layer of the force-informed residual policy is
initialized to zero. Residual training therefore starts from the
action mean produced by the pretrained imitator.

\section{Training Configuration}
\label{app:training}

\subsection{Simulation and PPO}
\label{app:ppo}

The main DexMachina and FoLD experiments use Genesis~\cite{genesis2024} for simulation and
PPO~\cite{schulman2017proximal} for policy optimization, implemented with
RL Games~\cite{makoviichuk2021rlgames}. Advantages are computed using
generalized advantage estimation (GAE)~\cite{schulman2015gae}. The simulator advances at
$60\,\mathrm{Hz}$, and the policy produces one action at every
simulation step. Each control step contains two physics substeps. The
shared PPO configuration is summarized in \autoref{tab:ppo_config}.

\begin{table}[h]
    \centering
    \caption{PPO and simulation configuration used in the DexMachina
    and FoLD experiments.}
    \label{tab:ppo_config}
    \begin{tabular}{lc}
        \toprule
        Parameter & Value \\
        \midrule
        Parallel environments                  & 4096 \\
        Rollout horizon                         & 16 \\
        Learning rate                           & $3\times10^{-4}$ \\
        PPO mini-epochs                         & 5 \\
        Minibatch size                          & 32768 \\
        Discount factor $\gamma$                & 0.99 \\
        GAE coefficient $\lambda_{\mathrm{GAE}}$& 0.95 \\
        PPO clipping parameter                  & 0.2 \\
        Maximum gradient norm                   & 1.0 \\
        Entropy coefficient                     & 0 \\
        Control frequency                       & $60\,\mathrm{Hz}$ \\
        Physics substeps                        & 2 \\
        \bottomrule
    \end{tabular}
\end{table}

We train one policy for each hand--task--method configuration.
Consequently, the reported DexMachina and FoLD results do not estimate
uncertainty over independent training seeds.

\subsection{Training Stages and Budgets}
\label{app:training_budgets}

The hand-motion imitator is trained independently for each hand--task
configuration for 1,000 epochs. The task-specific compensatory force-field encoder
is trained for 300 epochs and remains frozen during training of the
force-informed residual policy.

For both DexMachina and FoLD, clips containing at most 200 frames are
trained for at most 5,000 epochs. Longer clips are trained for at most
10,000 epochs. For these two clip-length groups, VOC assistance is
forcibly set to zero at epoch 4,000 and epoch 9,000, respectively, if
it has not already reached zero through the automatic curriculum.

If VOC assistance reaches zero earlier, policy training continues for
at most another 1,500 epochs, subject to the corresponding overall
training budget. If $e_0$ is the first zero-assistance epoch, the
stopping epoch satisfies

\begin{equation}
    e_{\mathrm{stop}}
    =
    \min(e_0+1500,e_{\max}),
\end{equation}

where $e_{\max}=5000$ for clips of at most 200 frames and
$e_{\max}=10000$ for longer clips.

\subsection{Reward Configuration}
\label{app:reward_config}

This subsection describes the full baseline policy training and FoLD's
stage-2 force-informed residual-policy training. In particular, the
action-penalty coefficient is 0.01 for both methods; these coefficients
do not describe stage-1 hand-imitator pretraining.

Following DexMachina~\cite{zhao2026dexmachina}, both methods use the same object-tracking, hand-imitation,
joint-reference, position-based contact, and control terms. FoLD
additionally uses a frame-to-frame hand-motion consistency term. The
compensatory force-field embedding itself is only a policy observation and has no
direct reward term.

The total reward is a weighted sum of the tracking and imitation terms,
minus the penalty terms, with coefficients in \autoref{tab:reward_weights}.
Object tracking combines exponential rewards for root position, root
orientation, and articulation. Hand-pose imitation compares wrist poses
and fingertip positions with the demonstration, while joint-reference
tracking compares robot joint positions with the retargeted reference.
The contact term matches contact-point locations for each object part.
FoLD's motion-consistency term compares consecutive-frame fingertip
displacements, wrist translations, and wrist rotation magnitudes with
their demonstrated counterparts. The high-contact-force penalty
penalizes contact-force magnitudes above a threshold, and the action
penalty is based on the mean squared action magnitude.

\begin{table}[t]
    \centering
    \caption{Reward coefficients for the DexMachina baseline and FoLD
    stage-2 residual-policy training.}
    \label{tab:reward_weights}
    \small
    \setlength{\tabcolsep}{3pt}
    \begin{tabular}{lcc}
        \toprule
        Term & DexMachina & FoLD \\
        \midrule
        Object tracking                   & 1.0  & 1.0 \\
        Hand-pose imitation               & 0.3  & 0.3 \\
        Frame-to-frame motion consistency & 0.0  & 0.1 \\
        Joint-reference tracking          & 0.3  & 0.3 \\
        Position-based contact tracking   & 3.0  & 3.0 \\
        High-contact-force penalty        & 0.1  & 0.1 \\
        Action penalty                    & 0.01 & 0.01 \\
        Direct force-field reward         & 0.0  & 0.0 \\
        \bottomrule
    \end{tabular}
\end{table}

The object-tracking reward uses exponential scales of 10, 1, and 5 for
root-position, root-rotation, and articulation errors, respectively.
The position-based contact reward uses a contact scale of 10.

\subsection{VOC Curriculum}
\label{app:voc}

The virtual object controller begins with position gain $k_p=80$,
velocity gain $k_v=5$, and force range 50. Curriculum statistics are
accumulated over a 30-episode deque after an initial waiting period of
100 epochs. A regular gain reduction requires

\begin{equation}
    \begin{aligned}
        \bar r_{\mathrm{task}}&\geq0.6, &
        \bar r_{\mathrm{con}}&\geq0.01,\\
        \bar r_{\mathrm{imi}}&\geq0.01, &
        \bar r_{\mathrm{bc}}&\geq0.01,
    \end{aligned}
\end{equation}

together with an average episode length of at least $T-2$, where $T$
is the clip length. Position and velocity gains use upper and lower
sampling-bound ratios of 0.9 and 0.8 after a successful curriculum
update. The force-range bound remains fixed during the regular
curriculum and is disabled together with the VOC at the forced
zero-assistance epoch.

The curriculum uses an episode-length rollback threshold of 80 and a
minimum rollback interval of 500 epochs. The forced zero-assistance
schedule in Sec.~\ref{app:training_budgets} guarantees that all selected
policies receive a period of training without VOC assistance.

\section{Evaluation Protocol}
\label{app:evaluation}

\subsection{Zero-Assistance Evaluation}
\label{app:zero_gain_eval}

All DexMachina, FoLD, and component-ablation policies are evaluated
with the virtual object controller completely disabled:

\begin{equation}
    k_p^{\mathrm{VOC}}=0,\qquad
    k_v^{\mathrm{VOC}}=0,\qquad
    f_{\mathrm{range}}^{\mathrm{VOC}}=0.
\end{equation}

Paired methods use identical clips, retargeted references, initial
states, physical parameters, and termination criteria. Environment
randomization is disabled, and deterministic policy actions are used.

Checkpoint selection is restricted to saved checkpoints obtained after
VOC assistance has reached zero. Among these checkpoints, we select
the checkpoint with the highest mean episode reward recorded during
training and subsequently evaluate it under the same zero-assistance
condition.

\subsection{Evaluation Metrics}
\label{app:metrics}

Let $T_i$ denote the number of evaluated post-action samples in rollout
$i$. A frame satisfies the object-tracking criterion when

\begin{equation}
    e^{\mathrm{pos}}_{i,t}\leq0.15\,\mathrm{m}
    \quad\text{and}\quad
    e^{\mathrm{rot}}_{i,t}\leq40^\circ.
\end{equation}

Let $L_i$ be the number of consecutive samples completed before the
first violation of either threshold. If no violation occurs, we set
$L_i=T_i$. We define the success rate used in our experiments as the
normalized completed trajectory length:

\begin{equation}
    \mathrm{SR}_i=\frac{L_i}{T_i}.
\end{equation}

A complete rollout therefore obtains $\mathrm{SR}=1$, whereas a rollout
that loses object tracking early receives a proportionally smaller
score.

Position error is the Euclidean distance between the achieved and
reference object-root positions. Rotation error is the geodesic angular
distance between the achieved and reference object-root orientations.
Joint error is the absolute difference between achieved and reference
object articulation angles. These errors are averaged over the
complete evaluated trajectory and reported in centimeters, degrees,
and degrees, respectively.

For the DexMachina comparison and component ablation, Contact Mean is
the temporal mean of the recorded position-based contact reward
$r_{\mathrm{con}}$. For the CHORD comparison, CWS Mean is the
temporal mean of CHORD's Contact Wrench Support (CWS) reward. The two
contact rewards use different definitions and numerical scales and are
therefore comparable only between methods within the same table.

\subsection{Macro-Averaging}
\label{app:macro_average}

For each hand, we first compute each metric independently on every task
and then take an unweighted macro-average over tasks. Short and long
clips therefore receive equal weight. The overall four-hand result is
an unweighted macro-average over all included hand--task cells rather
than an average weighted by clip length:

\begin{equation}
    \bar m_h
    =
    \frac{1}{|\mathcal D_h|}
    \sum_{d\in\mathcal D_h}m_{h,d},
    \qquad
    \bar m_{\mathrm{all}}
    =
    \frac{1}{|\mathcal V|}
    \sum_{(h,d)\in\mathcal V}m_{h,d},
\end{equation}

where $\mathcal D_h$ is the set of tasks evaluated for hand $h$, and
$\mathcal V$ is the set of included hand--task cells.

\section{Tasks and Demonstrations}
\label{app:tasks}

\subsection{DexMachina Comparison}
\label{app:dex_tasks}

The DexMachina comparison uses the same seven ARCTIC clips for Inspire,
Allegro, XHand, and Schunk. Within every hand--task cell, DexMachina
and FoLD share the same temporal range, retargeted reference, initial
state, and physical configuration. The complete task suite contains
1,670 reference frames per hand before exclusions.

\begin{table*}[t]
    \centering
    \caption{ARCTIC demonstrations used in the four-hand DexMachina
    comparison. Frame ranges use half-open intervals.}
    \label{tab:dex_task_list}
    \begin{tabular}{llllr}
        \toprule
        Task & Training identifier & ARCTIC sequence
             & Frame range & Length \\
        \midrule
        Ketchup-100
            & \texttt{ketchup-30-130-s01-use\_01}
            & \texttt{s01/use\_01} & $[30,130)$ & 100 \\
        Box-200
            & \texttt{box-30-230-s01-use\_01}
            & \texttt{s01/use\_01} & $[30,230)$ & 200 \\
        Mixer-170
            & \texttt{mixer-30-200-s01-use\_01}
            & \texttt{s01/use\_01} & $[30,200)$ & 170 \\
        Ketchup-300
            & \texttt{ketchup-30-330-s01-use\_01}
            & \texttt{s01/use\_01} & $[30,330)$ & 300 \\
        Mixer-300
            & \texttt{mixer-30-330-s01-use\_01}
            & \texttt{s01/use\_01} & $[30,330)$ & 300 \\
        Notebook-300
            & \texttt{notebook-30-330-s02-use\_01}
            & \texttt{s02/use\_01} & $[30,330)$ & 300 \\
        Waffleiron-300
            & \texttt{waffleiron-30-330-s01-use\_01}
            & \texttt{s01/use\_01} & $[30,330)$ & 300 \\
        \bottomrule
    \end{tabular}
\end{table*}

Ketchup-100 and Ketchup-300 are different temporal crops of the same
ARCTIC demonstration. Mixer-170 and Mixer-300 are defined similarly.
Box uses the original \texttt{para} retargeted reference, while all
remaining clips use the temporally repaired \texttt{temporal\_v1}
references.

The official DexMachina release does not provide complete frame-level
mappings for all reported clips. We therefore matched the Mixer and
Waffleiron sequences to the released qualitative videos according to
task identity and motion content. The same selected sequences are used
for both compared methods.

\subsection{Numerically Divergent Evaluation Case}
\label{app:divergent_case}

The complete evaluation suite contains 28 hand--task cells, comprising
seven tasks for each of four hand embodiments. On
Schunk--Ketchup-300, the DexMachina rollout underwent numerical
divergence. In the raw aggregate statistics, the object-position and
articulation errors reached approximately $10^{6}$ and $10^{12}$,
respectively. Directly including these values would dominate the
arithmetic means and make the aggregate tracking errors uninformative.

We therefore exclude Schunk--Ketchup-300 from the aggregate results.
To preserve a strictly paired comparison, the corresponding FoLD cell
is excluded as well. All metrics in
\autoref{tab:comparison_dexmachina} are consequently computed over
the same 27 hand--task cells. The exclusion is applied identically to
both methods.

\subsection{Component Ablation}
\label{app:ablation_tasks}

The component ablation uses the Inspire hand and three tasks:
Box-200, Mixer-170, and Notebook-147 (reference frames $[253,400)$). All variants share the same
task-specific motion-prior checkpoint, semantic residual architecture,
residual scale, reward configuration, and training budget. All variants
train a residual policy on top of the frozen imitator.

The three rows of \autoref{tab:ablation_pipeline} correspond to the
following configurations. VOC curriculum refers to assistance during
training, as described in Appendix~\ref{app:voc}.

\begin{itemize}
    \item The first row uses neither VOC training assistance nor
    compensatory force-field input.
    \item The second row uses VOC assistance during residual-policy
    training but omits the force-field input.
    \item The third row uses the same VOC curriculum and additionally
    provides the 128-dimensional compensatory force-field latent to the
    residual policy. This is FoLD as evaluated in the DexMachina comparison.
\end{itemize}

The reported results are unweighted macro-averages over the three
tasks. The comparison between the second and third rows isolates the
incremental contribution of compensatory force-field conditioning.

\section{CHORD Integration}
\label{app:chord_details}

CHORD and CHORD + FoLD use identical rewards, action mappings, VOC
schedules, optimization settings, training budgets, and evaluation
protocols. The only difference is that CHORD + FoLD appends the
task-specific 128-dimensional compensatory force-field embedding to the policy
observation and adjusts the corresponding network input dimension.
The integration retains CHORD's native residual control relative to
reference wrist and finger targets. The pretrained hand-motion imitator
used in the DexMachina-based FoLD pipeline is not used in this integration.

Both variants follow the original CHORD training configuration and are
trained for 12,000 epochs. Checkpoint selection is restricted to
checkpoints obtained after VOC assistance has reached zero. Among
these checkpoints, we select the checkpoint with the highest recorded
mean training reward.

The complete evaluation suite uses the five sequences listed in
\autoref{tab:chord_sequences}. No additional temporal cropping is
applied. The source demonstrations are recorded at $30\,\mathrm{Hz}$.
Evaluation runs at $20\,\mathrm{Hz}$ with
$\texttt{motion\_speed}=0.5$. The saved state trajectories include the
initial state, whereas rewards and tracking metrics are computed from
post-action samples. Each trajectory therefore contains one more saved
state than metric sample.

\begin{table}[t]
    \centering
    \caption{Complete sequences used in the CHORD comparison.}
    \label{tab:chord_sequences}
    \resizebox{\columnwidth}{!}{%
    \begin{tabular}{lrrr}
        \toprule
        Sequence
        & Source frames
        & Eval. states
        & Metric samples \\
        \midrule
        \texttt{s01\_box\_use\_01}
            & 889 & 1184 & 1183 \\
        \texttt{s07\_box\_grab\_01}
            & 725 & 965  & 964 \\
        \texttt{s01\_notebook\_use\_01}
            & 669 & 890  & 889 \\
        \texttt{s01\_laptop\_use\_01}
            & 653 & 869  & 868 \\
        \texttt{s01\_waffleiron\_use\_01}
            & 605 & 805  & 804 \\
        \bottomrule
    \end{tabular}
    }
\end{table}

For each method and sequence, we evaluate the same selected checkpoint
using five evaluation seeds. The results can differ across seeds. We
first average each metric over the five seeds for each sequence and
then compute an unweighted macro-average over all five sequences:
box-use, box-grab, notebook, laptop, and waffleiron. All five sequences
receive equal weight regardless of trajectory length. This aggregation
defines the comparison in \autoref{tab:comparison_chord}.

The complete CHORD evaluation contains

\begin{equation}
    5\ \text{sequences}
    \times
    2\ \text{methods}
    \times
    5\ \text{evaluation seeds}
    =
    50\ \text{rollouts}.
\end{equation}

The main table reports these means together with sample standard
deviations over the 25 task--rollout evaluations per method. These
standard deviations describe variation across tasks and evaluation
rollouts, not across independently trained policies; we do not claim
statistical significance.

\section{Qualitative Real-Robot Demonstrations}
\label{app:real_robot}

Our physical experiments use two hardware platforms. The first consists
of two Franka Research 3 (FR3) manipulators equipped with Sharpa Wave
hands. The second is a Galbot G3 equipped with two Galbot Sharpa hands.

The real-robot demonstrations are reconstructed from our in-house
videos. On the dual-FR3 platform, we demonstrate:

\begin{itemize}
    \item folding a pair of headphones; and
    \item closing a laptop screen.
\end{itemize}

On the Galbot G3 platform, we demonstrate headphone folding, laptop
closing, and opening and closing a pair of scissors.

For each task, the complete FoLD policy first generates a bimanual
robot trajectory in simulation. Before physical execution, we perform
an inverse-kinematics reachability check to determine a suitable
placement of the object. The generated trajectory itself is not
manually modified and is subsequently executed through open-loop
replay.

No online object-state feedback or policy inference is used during
physical execution. These demonstrations therefore evaluate the
physical executability of FoLD-generated motions rather than
closed-loop sim-to-real policy transfer. Since this evaluation is
intended as a qualitative proof of concept, we do not report a
task-level success rate.

\end{document}